# Decision language models as affordable and rapid alternatives to LLM judges of clinical generative AI

Authors: Rotem Gershon[1], Seffi Cohen[1,2], Dean Ariel[2,3], Havatzelet Bilavsky-Yarden[4,5], Amiel Sberro[4,5], Shai Yitzhaki[4,5], Osnat Tausky[4,5], Naama Golan[4,5], Maayan Rabinovitch[4,5], Ariel Avraham Hasidim[4,5], Itay Arianu[4,5], Maya Dagan[1], Ran Balicer[1,6,7], Noa Dagan[1,2,6]

[1]Innovation Division, Clalit Research Institute, Clalit Health Services, Ramat-Gan, Israel. [2]Stein Faculty of Computer and Information Science, Ben Gurion University of the Negev, Be'er Sheva, Israel. [3]Department of Family Medicine, Clalit Health Services, Tel Aviv, Israel. [4]Schneider Children's Medical Center, Petah-Tikva, Israel. [5]Gray Faculty of Medical & Health Sciences, Tel Aviv University, Tel Aviv, Israel. [6]The Ivan and Francesca Berkowitz Family Living Laboratory Collaboration at Harvard Medical School and Clalit Research Institute, Ramat-Gan, Israel. [7]Faculty of Health Sciences, School of Public Health, Ben Gurion University of the Negev, Be'er Sheva, Israel.

Corresponding author: Rotem Gershon (rotem.gershon@gmail.com).

**Abstract**

Automated judges of clinical artificial intelligence (AI) outputs are necessary, but large language model (LLM) judges are costly and slow at scale. We compared Jev, a decision language model that returns structured judgments, with frontier and open-weight LLM judges in four use cases. On 118 pediatric answers, Jev's agreement with physicians' ordinal quality ratings (Gwet's AC2 0.50; expected score 0.53) matched or exceeded that of seven of 12 LLM judges, including frontier models, Claude Opus 5.5 (0.49) and GPT-6 (0.38). On physician-graded binary HealthBench criteria, Jev matched or exceeded four of six LLM judges. On MedHELM's open-ended tasks, Jev's agreement with a three-LLM jury approached or exceeded the jurors' internal agreement. On HealthBench Professional, Jev matched the benchmark's reference judge, GPT-5.4, on 91% of criteria (other judges, 91–93%). Frontier judges cost 73–294 times more than Jev and took 7–57 times longer. Jev may enable efficient clinical AI evaluation at scale.

The output of clinical generative artificial intelligence (AI) requires continuous evaluation, because each model update, prompt change, or knowledge-base revision can change the quality and safety of its answers[1,2]. Physician grading is the reference standard[3], but its cost and time limit how often it can be repeated.

To fill this gap, large language models (LLMs) have been used to grade the output of other models. This LLM-as-a-judge approach comes close to inter-human agreement in general-domain benchmarks, despite documented weaknesses[4–10]. In medicine, LLM judges have evaluated health conversations, clinical summaries and answers, and MedHELM's open-ended tasks, with selected judges achieving agreement with clinicians comparable to inter-clinician agreement[11–15]. Whether this extended to a deployed clinical decision-support system across distinct quality and safety criteria remained unclear. We therefore recently compared seven LLM judges with five physicians on 118 answers from such a system in a tertiary pediatric hospital[16]. LLM judges agreed more closely with physicians on text-comparison rubrics, such as faithfulness and completeness, than on reasoning-heavy rubrics, such as medical reasoning and safety.

Despite the LLM judges' promising performance on some evaluation tasks, they are costly: frontier and reasoning models are priced per generated token, and a judge that writes a rationale or hidden chain of thought before each score can consume thousands of tokens per judgment[17,18]. When the system's performance needs to be evaluated on full test sets, this cost is multiplied by several orders of magnitude. We borrow the oncology term financial toxicity[19] for the recurring, unpredictable expense that token-priced models might impose on health systems[20]. Token-by-token generation is also slow, and when a check runs in real time while a clinician waits, a delay of several seconds may affect usability and adoption[21].

Autoregressive LLMs produce their judgments token by token, although a judge needs to return only a score or a pass or fail decision. Jev (TypeSafe AI) is a non-autoregressive decision language model that produces its outputs in parallel rather than token by token[22,23]. According to its developers, it evaluates all questions about an input in parallel and returns, instead of text, a categorical choice or an ordinal score, each with a confidence value, or the probability of yes to a binary question[24,25]. Its architecture and size are not disclosed, and its list price is currently US$0.042 per million input tokens, with no charge for output[25].

Jev is not meant to compete with LLMs on all tasks; Jev's developers document weaknesses relevant to clinical grading, among them reduced accuracy on questions that require multiple hops of reasoning, sensitivity to irrelevant detail in the input and literal reading of instructions[26]. They advise providing Jev with structured tasks[25]. They also suggest that because Jev is trained to return calibrated probabilities, the user can use only high-confidence outputs[25,27]. Indeed, when Jev was evaluated on such structured clinical tasks, it provided useful outputs, especially for cases in which the output received a confidence score of 100%[28].

Evaluating the quality of AI system outputs is a relatively structured task that often depends largely on the context provided for evaluation rather than extensive general knowledge. It must be repeated across entire test sets whenever a system changes during development, and sometimes performed rapidly on individual outputs in real time. These characteristics suggest that decision models such as Jev could be well suited to the task: its structure may support accurate judgments, while their low cost and latency may enable evaluation at scale and in real time.

Here we compared Jev with LLM judges, frontier, low-cost and open-weight, in four AI output evaluation tasks; the evaluation was compared against a reference standard of grades given by physicians in two use cases and by LLMs in two others. The first task included the 118 pediatric answers of our earlier study[16], rated on ordinal and binary scales. The second used the HealthBench meta-evaluation set, in which physicians made only binary judgments[11]. The third evaluation used the MedHELM benchmark, whose open-ended tasks are graded by a jury of three LLMs that its authors validated against clinician ratings[15]. The fourth is HealthBench Professional, a published 2026 benchmark of clinicians' tasks[29] whose original grader is GPT-5.4. Agreement with these graders tests whether Jev could replace them in established benchmarks. We also measured Jev's cost and speed against those of the LLM judges and tested a hybrid approach in which an LLM judge grades the judgments Jev defers. Finally, we examined whether Jev shared the LLM judges' weaknesses.

## Results

### Stability of Jev's ratings across repeated runs in the four use cases

Jev graded every use case three times, and stability was the Gwet's Agreement Coefficient 2 (AC2) between runs (use cases 1 and 3) or the share of identical labels (use cases 2 and 4). Jev's ratings were stable across three runs in all four use cases (Extended Data Table 1, Fig. 1): AC2 agreement between runs was 0.99–1.00 in use cases 1 and 3 with either reading of Jev's scores (Methods), compared to 0.89–1.00 for the 12 LLM judges in use case 1, and labels were identical in all three runs for 97.7% and 98.8% of rows in use cases 2 and 4, respectively.

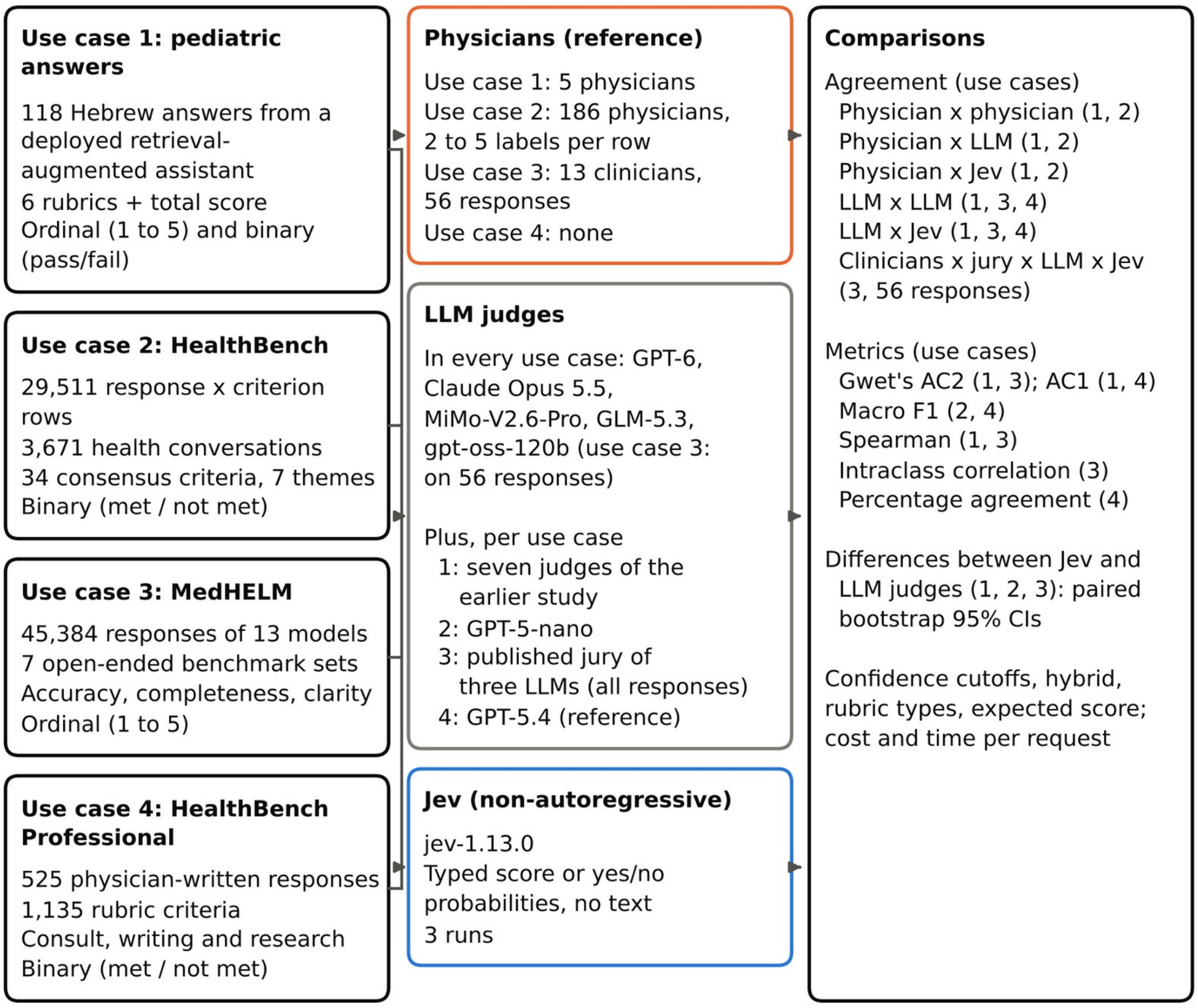

**Fig. 1 | Study design.** Four use cases were graded by autoregressive large language model (LLM) judges and by Jev, a non-autoregressive decision language model. Use case 1 comprised 118 answers from a retrieval-augmented pediatric clinical decision-support system, rated by five physicians and 12 LLM judges (the seven of our earlier study, GPT-6, Claude Opus 5.5, MiMo-V2.6-Pro, GLM-5.3 and this study's run of gpt-oss-120b) on six rubrics and a total score on an ordinal (1 to 5) scale, and nine of the LLM judges (four published, this study's five) on the six rubrics on a binary (pass or fail) scale. Use case 2 comprised the HealthBench meta-evaluation set, in which physicians made binary judgments of whether model responses to health conversations met rubric criteria, also graded by GPT-5-nano, GPT-6, Claude Opus 5.5, MiMo-V2.6-Pro, GLM-5.3 and gpt-oss-120b. Use case 3 comprised 45,384 responses of 13 models to the open-ended MedHELM benchmarks, rated on a 5-point Likert scale for accuracy, completeness and clarity by a published jury of three LLMs and, on 56 clinician-rated responses, by GPT-6, Claude Opus 5.5, MiMo-V2.6-Pro, GLM-5.3 and gpt-oss-120b. Use case 4 comprised 525 physician-written HealthBench Professional responses graded against 1,135 rubric criteria by GPT-5.4, the benchmark's default grader, and by the five LLM judges of use case 2 other than GPT-5-nano. Use case 4 has no physician labels, and use case 3 has clinician ratings for 56 responses. Agreement was compared between pairs of physicians, between physicians and LLM judges, between physicians and Jev, between LLM judges and between LLM judges and Jev (use cases 1, 3 and 4), and, in use case 3, between clinicians, the jury, these judges and Jev. Numbers in brackets in the Comparisons box name the use cases.

## Use case 1: agreement of Jev and the LLM judges with physicians on 118 pediatric answers

Jev and 12 LLM judges, the seven of our earlier study[16] and the five LLMs chosen for this study (GPT-6, Claude Opus 5.5, MiMo-V2.6-Pro, GLM-5.3 and gpt-oss-120b), rated 118 pediatric decision-support answers on six rubrics and a total score from 1 to 5. Each score was compared with the physician panel (median of five) by Gwet's AC2. Averaged over the six rubrics and the total score, Jev's agreement with physicians was AC2 0.50 for its most probable level and 0.53 for its expected score, the sum of its levels weighted by their probabilities (95% CI 0.44 to 0.56 and 0.49 to 0.58; Methods, Extended Data Table 2, Fig. 2a). With either reading (most probable level or expected score), it matched Claude Opus 5.5 (0.49), MiMo-V2.6-Pro (0.50), the published run of gpt-oss-120b (0.50) and o3 (0.55), and exceeded GPT-6 (0.38), Claude Sonnet 4.5 (0.41) and GPT-5 (0.44). Five judges agreed more closely (0.60 to 0.81), led by Llama 4, GPT-4o and GLM-5.3 (Extended Data Table 2 note). Jev agreed with the 12 LLM judges more closely than they agreed with one another (mean AC2 0.86 versus 0.82 with either reading; Fig. 3a, Supplementary Table 1). By rubric, with either reading, Jev agreed with physicians more closely than the mean of the seven published judges on safety (0.78 versus 0.71) and less closely on faithfulness, actionability and medical reasoning; its expected score also agreed more closely on conciseness and less closely on completeness (Extended Data Table 2, Figs. 2a and 3c). Agreement on the binary pass or fail scale is shown in Fig. 2b, and the detection of the answers physicians failed, by rubric, in Supplementary Table 2.

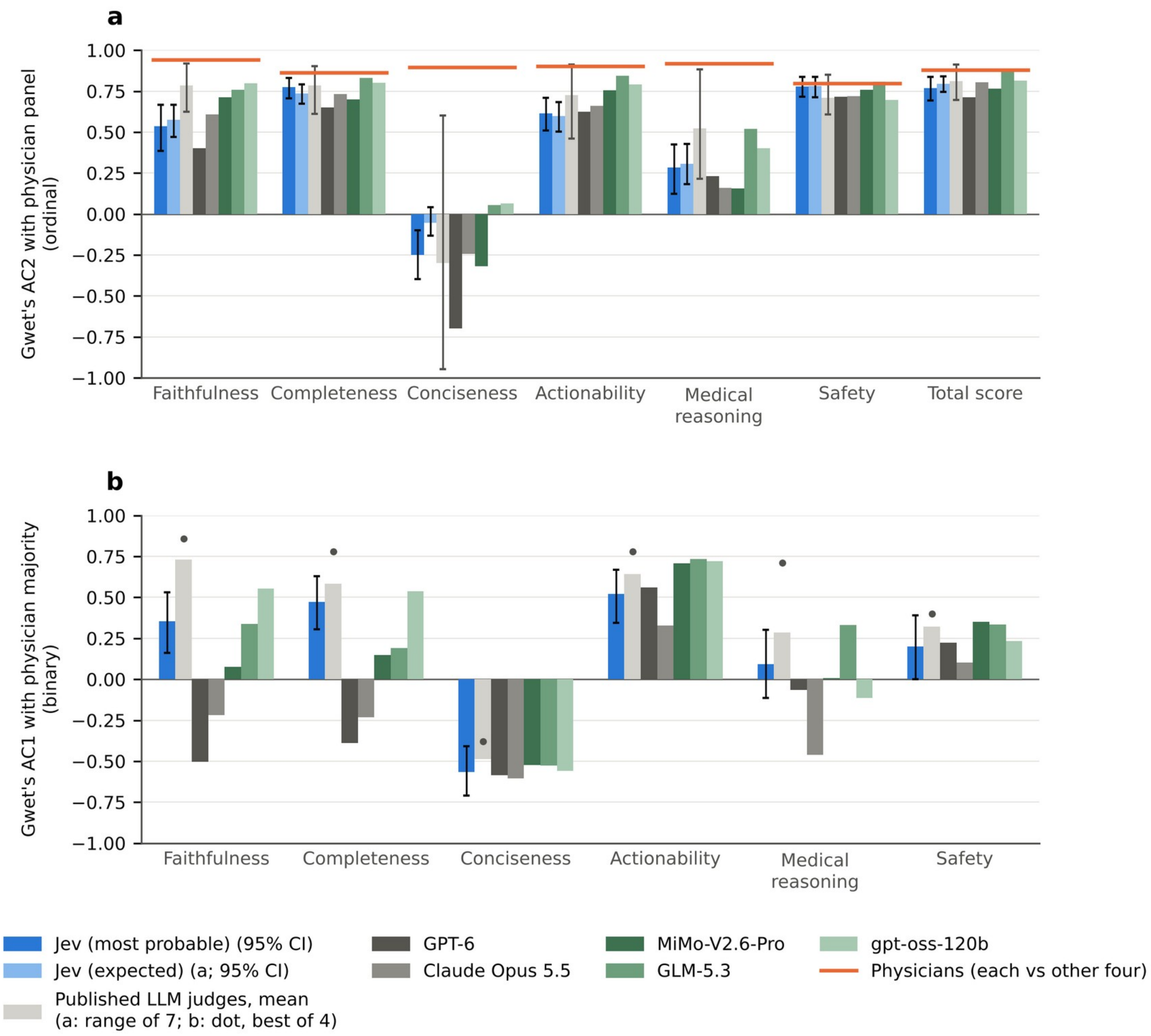

**Fig. 2 | Agreement with physicians by rubric in use case 1. a**, Gwet's AC2 (quadratic weights) between each judge's ordinal ratings and the physician panel (median of five physicians), first run. Bars, from left: Jev (most probable) and Jev (expected), its expected score rounded half-up (Methods), each with bootstrapped 95% confidence interval (2,000 resamples of answers); the mean of the seven published LLM judges, with their range; GPT-6; Claude Opus 5.5; MiMo-V2.6-Pro; GLM-5.3; and this study's run of gpt-oss-120b. Orange lines, mean agreement of each physician with the panel of the other four. **b**, Gwet's AC1 between binary ratings and the physician majority (physician ratings of 4 or 5 counted as pass), for the same judges (GLM-5.3, 115 answers); the published bar is the mean of four LLM judges and the dot their best, as published previously, without confidence intervals. In use case 1 the physicians' binary reference is derived from their ordinal ratings, whereas judges answered a separate pass or fail prompt.

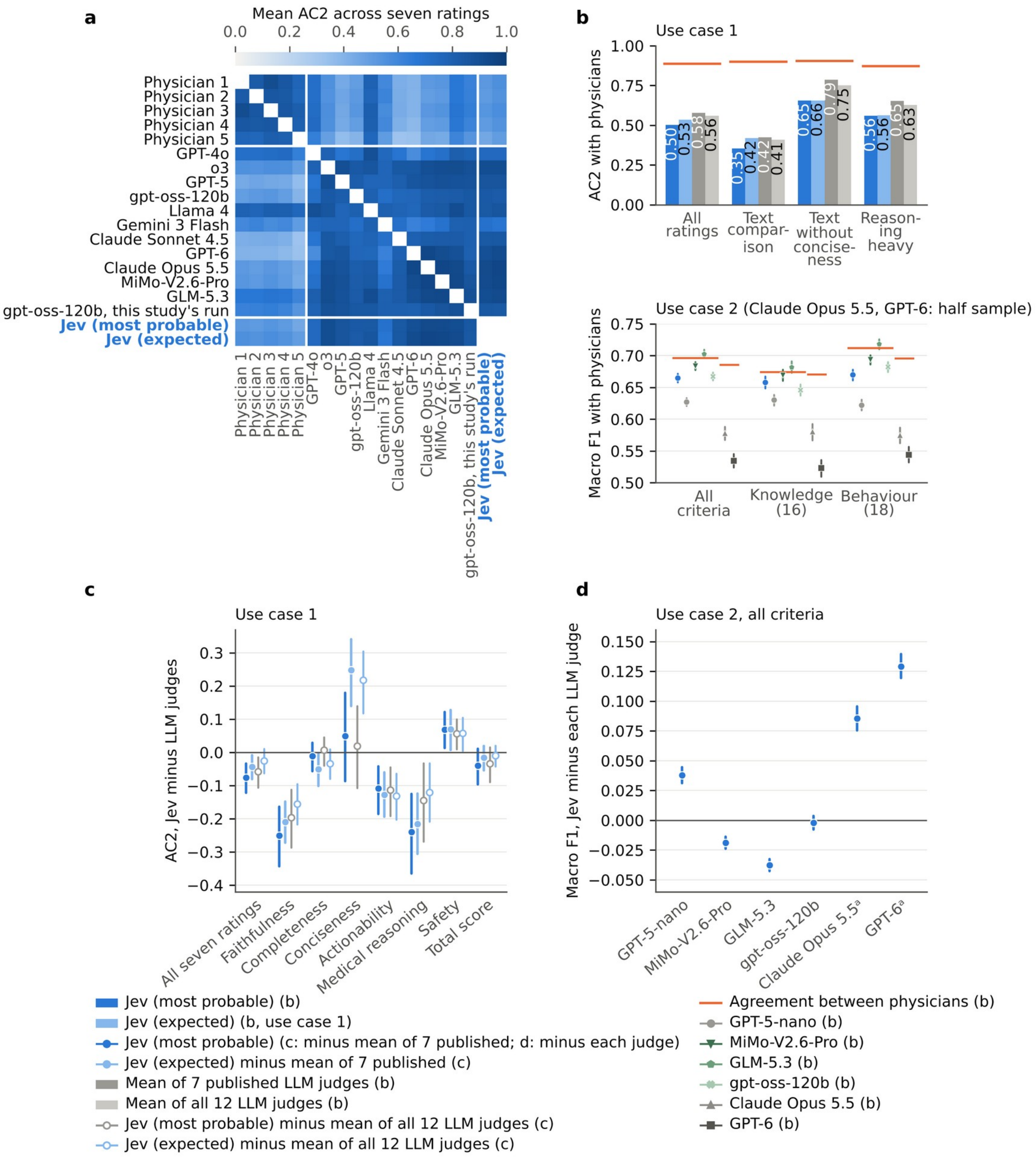

**Fig. 3 | Agreement with physicians in use cases 1 and 2: between raters, by rubric type, and differences between Jev and the LLM judges.** **a**, Pairwise Gwet's AC2 in use case 1, averaged over seven ratings (first run for LLM judges and Jev). White lines separate physicians, LLM judges and Jev; Jev (expected), as in Fig. 2 (its agreement with Jev (most probable) not computed). Values in Supplementary Table 1. **b**, Agreement with physicians overall and by rubric group; orange lines, agreement between physicians. Use case 1 (AC2, first run): Jev (most probable), Jev (expected), the mean of the seven published judges and the mean of all 12 LLM judges; physicians, each against the median of the other four; text-comparison rubrics (faithfulness, completeness, conciseness), the same group without conciseness, and reasoning-heavy rubrics (actionability, medical reasoning, safety). Use case 2 (macro F1 with 95% confidence intervals): Jev, GPT-5-nano and the open-weight judges on all rows (GLM-5.3, 29,507), and Claude Opus 5.5 and GPT-6 on the half sample, with physicians on the same rows; knowledge criteria (16 criteria on medical accuracy, safety or appropriateness) and behaviour criteria (18 criteria on context seeking, hedging, tailoring, level of detail and instruction following), listed in Supplementary Table 6; values in Extended Data Table 4. **c**, **d**, Jev's agreement with physicians minus the comparator's, with 95% paired bootstrap confidence intervals; grey line, no difference. **c**, Use case 1: AC2 of Jev (most probable) and Jev (expected) minus the mean of the seven published LLM judges (filled; blue and light blue) and minus the mean of all 12 judges (open; grey and light blue), averaged over seven ratings and for each rubric (2,000 resamples of answers). **d**, Use case 2: macro F1 of Jev minus each LLM judge on all criteria ([a]Claude Opus 5.5 and GPT-6 on the 14,762-row half sample; 2,000 resamples of conversations); results by criteria group and theme are in Extended Data Table 4.

### Use case 2: agreement of Jev and the LLM judges with physicians on HealthBench

In HealthBench, 186 physicians judged whether model responses met rubric criteria in 29,511 pairs of a response and a criterion[11]. Jev and four LLM judges judged every pair, Claude Opus 5.5 and GPT-6 a half sample, and each was compared with the physicians' labels by the HealthBench macro F1. Jev's macro F1 with physicians was 0.665, level with gpt-oss-120b (0.667), and above GPT-5-nano (0.627); on the half sample stratified by theme, which covered all 34 criteria, it was 0.664, against 0.578 for Claude Opus 5.5 and 0.535 for GPT-6 (Table 1, Fig. 3d). MiMo-V2.6-Pro (0.684) and GLM-5.3 (0.703) agreed more closely, as did the published HealthBench graders GPT-4.1, o4-mini and o3 (0.681 to 0.709)[11].

## Table 1 | Agreement, cost and time of Jev and the LLM judges in use cases 1, 2 and 4

| Judge | Use case 1: AC2 with physicians | Use case 2: macro F1 with physicians | Use case 4: agreement with GPT-5.4 | US$ per million judgments, use cases 1 / 2 / 4 | Median time per request (s), use cases 1 / 2 / 4 |
|---|---|---|---|---|---|
| Jev | 0.50[g] (0.44 to 0.56); Jev (expected) 0.53 (0.49 to 0.58) | 0.665 (0.658 to 0.672); half sample 0.664 (0.654 to 0.673) | 90.8% (89.0 to 92.4) | 38 / 54 / 89 | 0.55 / 0.12 / 0.51 |
| 7 published judges | 0.41 to 0.81 | – | – | 19.5 to 1,582[d] / – / – | – |
| GPT-5.4 | – | – | Reference | – / – / 6,556 | – / – / 3.52 |
| GPT-5-nano | – | 0.627 (0.621 to 0.634) | – | – / 273 / – | – / 9.32[e] / – |
| Claude Opus 5.5 | 0.49 (0.44 to 0.54) | 0.578[a] (0.568 to 0.588) | 92.0%[b] (90.3 to 93.5) | 4,854 / 10,407 / 17,047 | 5.72 / 6.75 / 3.14 |
| GPT-6 | 0.38 (0.31 to 0.45) | 0.535[a] (0.524 to 0.545) | 93.3% (92.0 to 94.7) | 6,490 / 11,007 / 21,083 | 6.07 / 4.38 / 2.94 |
| MiMo-V2.6-Pro | 0.50 (0.46 to 0.55) | 0.684 (0.678 to 0.690) | 91.6% (90.0 to 93.1) | 326 / 1,214 / 1,016 | 17.29 / 13.29 / 7.28 |
| GLM-5.3 | 0.67 (0.64 to 0.71) | 0.703[c] (0.696 to 0.709) | 91.3% (89.5 to 93.0) | 1,624 / 8,837 / 6,854 | 43.68 / 22.41 / 7.00 |
| gpt-oss-120b | 0.62[f] (0.57 to 0.67) | 0.667 (0.661 to 0.673) | 90.9% (89.3 to 92.6) | 42 / 137 / 127 | 18.33 / 11.68 / 9.89 |

In brackets, 95% CI (2,000 bootstrap resamples of answers in use case 1 and of conversations in use case 2; 1,000 of criteria in use case 4); differences between Jev and each judge are in Extended Data Table 2's note and Fig. 3d. [a]Half sample of the use case 2 conversations, stratified by theme and covering all 34 criteria: the 14,762 rows in 1,837 conversations labelled by Claude Opus 5.5, GPT-6 and GPT-5-nano; on these rows Jev 0.664, GPT-5-nano 0.628, MiMo-V2.6-Pro 0.686, GLM-5.3 0.701, gpt-oss-120b 0.666 and agreement between physicians 0.686. [b]1,121 criteria; 13 outputs put the JSON object after a prose preamble and 1 was a refusal. [c]29,507 rows; 4 rows without a label (2 exceeded the output limit, 2 not completed). [d]Ordinal prompt only (three runs, 2,478 judgments; the same per million as the first run); input tokens only, estimated (Extended Data Table 3). [e]Sequential requests. [f]This study's run; the original study's run gave 0.50. [g]Jev (most probable), followed by Jev (expected), the two readings of Jev's scores (Methods); use cases 2 and 4 have one reading. Use case 1: mean of the seven ratings, first run; use case 2: all 29,511 rows unless marked; use case 4: 1,135 criteria. Costs at the prices billed: October 2026 list prices (batch prices for Claude Opus 5.5 and GPT-6 apart from their synchronous rows, and for GPT-5-nano on all rows although 10,021 ran synchronously, in use case 2) or as billed by OpenRouter for the open-weight models, including provider cache discounts; costs at standard list prices are in Supplementary Table 7. Costs per million judgments as in Fig. 5: use case 1, first ordinal run (826 judgments), per rubric rating as for Jev; use case 2, all 29,511 rows sent or the 14,762 half-sample rows compared; Jev's use case 2 agreement uses the majority label of three runs. Time per request: use case 1, ordinal prompt, six concurrent requests (Claude Opus 5.5 and GPT-6 with their six runs in parallel), LLM judges pooled over three runs and Jev from a separate timing pass of 118 requests; use case 2, the same 150 rows at six concurrent requests (Jev's sequential median, 0.16 s, in Extended Data Table 3); use case 4, 16 concurrent requests for Jev and GPT-5.4 and six for the other judges. Use case 3 is graded by its published jury (Extended Data Table 3, Fig. 5) and, on 56 clinician-rated responses, by GPT-6, Claude Opus 5.5, MiMo-V2.6-Pro, GLM-5.3 and gpt-oss-120b. AC2, Gwet's agreement coefficient 2.

## Use case 3: agreement of Jev with clinicians and the MedHELM LLM jury

### Agreement of Jev, the MedHELM LLM jury and the five LLM judges with clinicians on 56 responses

In MedHELM, a jury of three LLMs grades the responses of 13 LLMs to open-ended medical tasks; its authors validated the jury against clinician ratings of 56 responses from ACI-Bench[30] and MEDIQA[31] (intraclass correlation 0.47, against 0.43 between clinicians)[15]. Jev, the original benchmark's jury and the five LLM judges rated these responses on accuracy, completeness and clarity, and each rater's mean over the three was compared with the clinicians' mean by the authors' consistency intraclass correlation, ICC(3,2). On these responses, Jev's expected score agreed with the clinicians as closely as the jury (intraclass correlation 0.49 versus 0.47), and its most probable level less closely (0.31; Fig. 4a, Supplementary Table 3). Jev's expected score ranked third of the seven graders, the jury and the five LLM judges, after GLM-5.3 and Claude Opus 5.5 (0.52 each); none differed detectably from either reading of Jev (Fig. 4a, Supplementary Table 3).

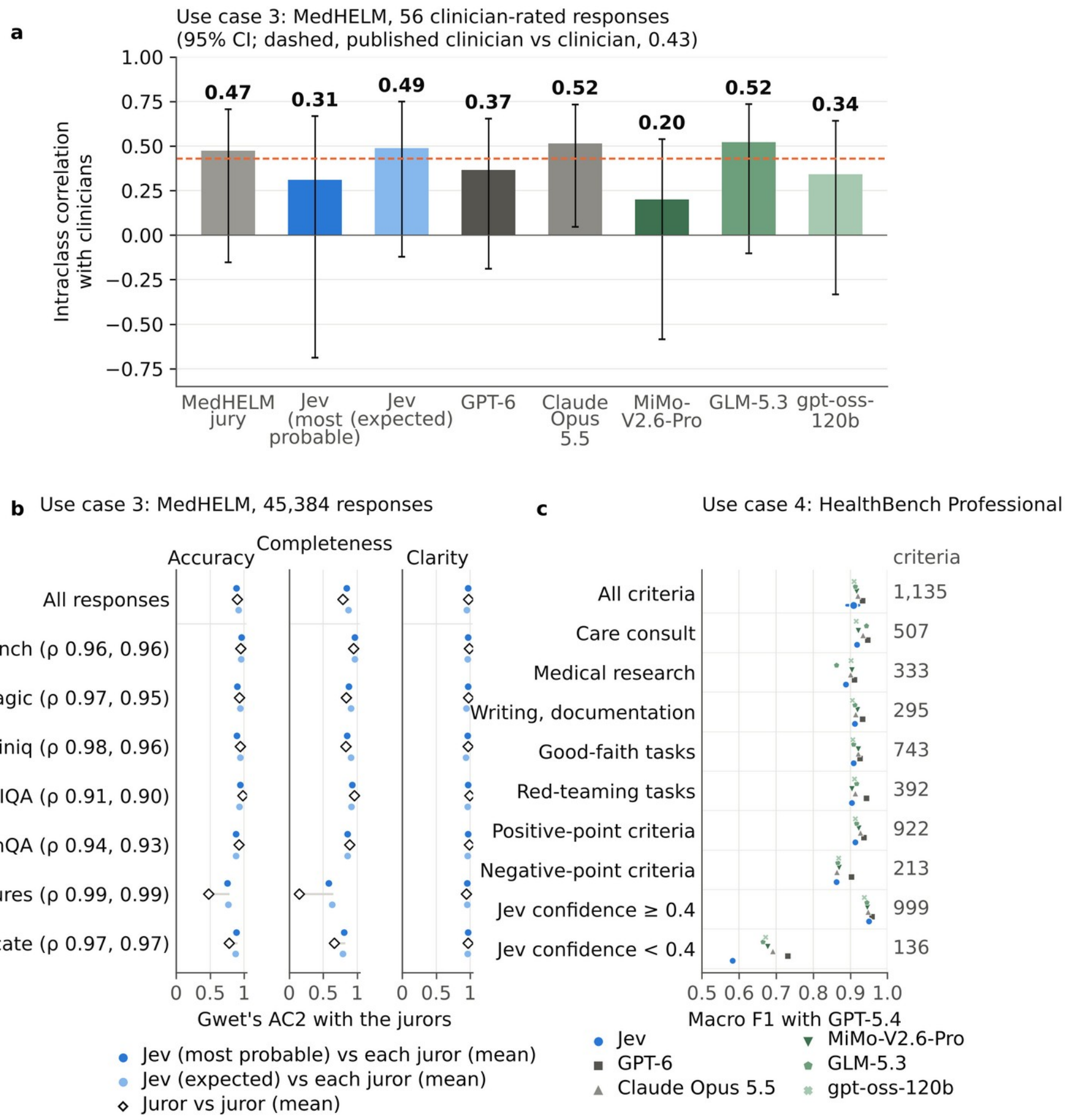

**Fig. 4 | Agreement of Jev and the LLM judges with clinicians and with the reference LLM graders in use cases 3 and 4. a**, Use case 3 (MedHELM), 56 clinician-rated responses from ACI-Bench and MEDIQA: consistency intraclass correlation ICC(3,2) between each rater's standardized score (mean of accuracy, completeness and clarity) and the clinicians' mean, with 95% confidence intervals (2,000 bootstrap resamples of responses), for the MedHELM jury, Jev (most probable), Jev (expected; unrounded, Methods) and the five LLM judges; dashed line, published agreement between clinicians (0.43)[15]. Values in Supplementary Table 3. **b**, Use case 3, all 45,384 responses: Gwet's AC2 between Jev and each of the three jurors (GPT-4o, Claude 3.7 Sonnet, Llama 3.3 70B), averaged over jurors (Jev (most probable), blue, above; Jev (expected), light blue, below), and between pairs of jurors, averaged over pairs (open diamonds), for each axis, overall and by benchmark set; ρ, Spearman correlation across the 13 evaluated models between their mean rating from Jev (most probable) and from Jev (expected) and their mean jury score, both averaged over the three axes; the five LLM judges graded only the 56 clinician-rated responses (a). Values in Supplementary Table 3. **c**, Use case 4 (HealthBench Professional): macro F1 with GPT-5.4 (reference) on 1,135 criteria for Jev (blue; overall with bootstrapped 95% confidence interval, 1,000 resamples of criteria) and the five other LLM judges (Claude Opus 5.5, 1,121 criteria), overall and by subset; the last two rows split the criteria by Jev's confidence $|2p - 1|$ at 0.4. Numbers on the right, criteria per subset. Values in Supplementary Table 4.

## Agreement of Jev with each juror of the MedHELM LLM jury on 45,384 responses

Jev also rated the 45,384 responses on the jury's 5-point scale. Treated as a fourth juror, Jev agreed with the three jurors nearly as closely as they agreed with one another, and more closely on completeness (mean AC2 0.84 for its most probable level and 0.87 for its expected score, versus 0.79); its expected score also agreed more closely on accuracy (0.92 versus 0.90). With either reading, Jev ranked the 13 evaluated models in nearly the same order as the jury (Spearman 0.91 to 0.99 for its most probable level and 0.90 to 0.99 for its expected score across the seven benchmark sets; Fig. 4b, Supplementary Table 3).

## Use case 4: agreement of Jev with GPT-5.4 on HealthBench Professional

In HealthBench Professional, responses to real clinician tasks are graded against physician-written criteria[29]. The release has no physician labels, so Jev and the five LLM judges judged whether each criterion was met and were compared with GPT-5.4, the benchmark's default grader. On its 525 physician-written responses, Jev and GPT-5.4 gave the same label on 90.8% of 1,135 criteria (95% CI 89.0% to 92.4%; macro F1 in Fig. 4c), level with gpt-oss-120b (90.9%) and within 2.5 percentage points of every other judge (Table 1, Supplementary Table 4).

## Cost and time of Jev and LLM judges across the four use cases

Cost came from billed tokens (estimated for the MedHELM jury from its published prompts and outputs), and time was the client's wall-clock time per request. Per million judgments at the prices billed, Jev cost US$22 to US$89 across the use cases; on the same judgments, the frontier judges cost 73 to 294 times as much, the open-weight judges 1.1 to 163 times, and the MedHELM jury an estimated 184 times (Fig. 5, Extended Data Table 3). The open-weight judges ran through a hosted application programming interface at the prices billed; hosting them on premises has hardware and operating costs, which were not measured. Jev's median time per request was 0.12 to 0.55 s. Measured under comparable conditions (use cases 1 and 2, and GPT-5.4 in use case 4), the frontier judges took 7 to 57 times as long per judgment and the open-weight judges 31 to 190 times (Table 1 and Extended Data Table 3).

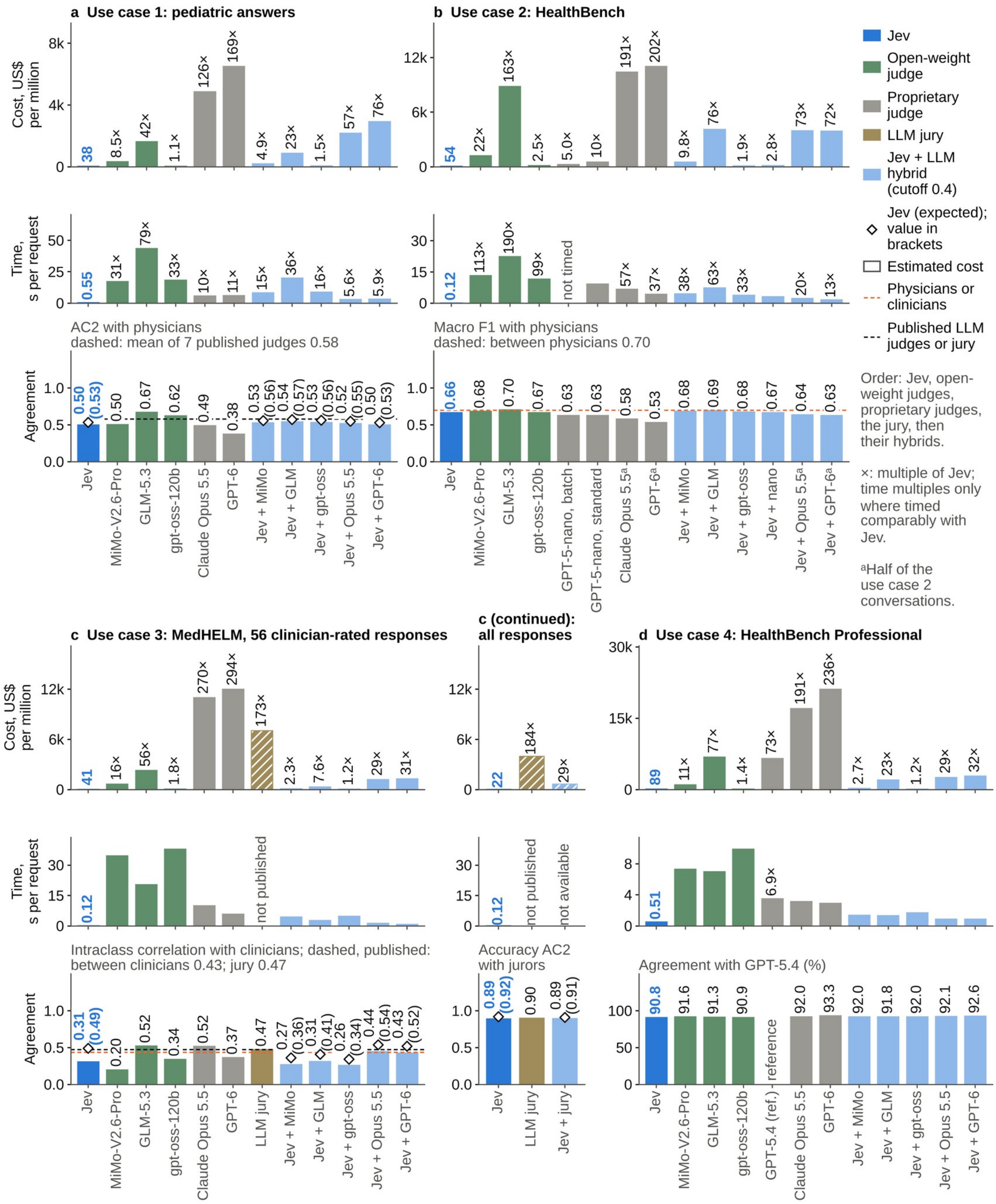

**Fig. 5 | Cost, time and agreement of Jev, LLM judges and Jev and LLM hybrids, by use case.** Panels a to d, use cases 1 to 4 (c, use case 3 in two blocks: 56 clinician-rated responses, all responses); in each, top, cost of one million rubric judgments at the prices billed (GPT-5-nano at batch and at standard prices); middle, median time per request; bottom, agreement with the reference. Bars run Jev, open-weight judges, proprietary judges, the jury, then their hybrids (Jev + judge); multiples of Jev end in ×. Each LLM judge is shown with its hybrid at the 0.4 cutoff (other cutoffs except the jury's, Extended Data Fig. 2); a hybrid's cost includes Jev on every judgment, and its time is Jev's plus the LLM judge's for the deferred share. Costs are on the same judgments within a block except the use case 2 half sample (Table 1 and Extended Data Table 3). Hatched bars are estimates: the jury's cost, from its published prompts and juror outputs at list prices (Supplementary Table 7), and its hybrid's, with one juror grading the 46% of responses with at least one judgment below the cutoff, averaged over the jurors. Time multiples appear only where judges were timed comparably: use case 1, ordinal prompt, LLM judges over three runs; use case 2, the same 150 rows at six concurrent requests (GPT-5-nano and its hybrid sequentially, without multiple); use case 4, Jev and GPT-5.4 at 16 concurrent requests. Agreement, from Jev's first run, for Jev (most probable) (bars) and Jev (expected) and the hybrids built on it (diamonds, values in brackets; use cases 1 and 3): AC2 with the physician panel, mean of seven ratings (use case 1); macro F1 with physicians (use case 2; ᵃClaude Opus 5.5 and GPT-6 on the half sample); intraclass correlation with clinicians (use case 3, 56 responses); accuracy AC2 with the jurors, averaged over jurors, juror pairs or single-juror fallbacks (use case 3, all responses); and agreement with GPT-5.4 (use case 4). Dashed lines, mean of the seven published judges (use case 1), agreement between physicians (use case 2) and the published intraclass correlations between clinicians (0.43) and of the jury (0.47)[15] (use case 3).

### Agreement after deferring Jev's low-confidence judgments

Jev kept only the judgments on which its confidence reached a cutoff, 0.5 for its score confidence in use cases 1 and 3, the developers' documented floor, and 0.4 for $|2p - 1|$, for a yes probability p, in use cases 2 and 4, and every rater was compared on the kept judgments. Setting aside low-confidence judgments, as Jev's developers recommend[27], raised Jev's agreement in all four use cases by small absolute amounts, and most of the other raters also agreed more closely on the kept judgments (Extended Data Fig. 1 for use cases 1, 2 and 4; Supplementary Tables 4 and 5). Jev kept 68% of use case 1 judgments, on which its AC2 was 0.57 for its most probable level and 0.60 for its expected score (0.50 and 0.53 on all judgments), against 0.60 for the mean of the seven published judges (Supplementary Table 5). On the 67% of use case 2 judgments it kept, Jev's macro F1 was 0.735 (0.665 on all), against 0.725 between physicians on the same rows (Supplementary Table 5), and on 88% of use case 4 criteria, it agreed with GPT-5.4 on 95.0% (90.8% on all; Supplementary Table 4).

### Performance-cost trade-off of hybrid evaluation by Jev and an LLM judge

The hybrid is a proposed way to keep costs low while letting a more capable LLM judge decide where Jev is least confident (Methods). To test its utility, Jev's judgments below a confidence cutoff of 0.4 were replaced by an LLM judge's. The hybrids added at most 0.04 to Jev's agreement in use cases 1 and 2 and 1.0 to 1.8 percentage points in use case 4, at 1.2 to 76 times Jev's cost. In use case 1, routing only the safety rubric to Claude Opus 5.5 or GPT-6 at the 0.2 cutoff raised the sensitivity for answers failed by physicians on safety from Jev's 31% to 49% and 51%, respectively (Supplementary Table 2). Conversely, with the same models, the hybrids lowered agreement in use case 2 (Fig. 5, Extended Data Fig. 2). In use case 3, the jury hybrid added less than 0.01 to the accuracy agreement of Jev's most probable level (0.893 versus 0.888) and lowered that of its expected score (0.908 versus 0.917). On the 56 clinician-rated responses of use case 3, no hybrid agreed detectably better than Jev's expected score alone (0.49, against 0.54 and 0.52 for the expected-score hybrids with Claude Opus 5.5 and GPT-6), and the expected-score hybrids with the other three judges agreed detectably less well (by 0.08 to 0.15).

### Jev's agreement on text-comparison versus reasoning-heavy rubrics

To test whether Jev's agreement with physicians differed between text-comparison rubrics (faithfulness, completeness, conciseness) and reasoning-heavy ones (actionability, medical reasoning, safety), as LLM judges' agreement does[16], its agreement was compared between these groups and, in use case 2, between knowledge (16) and behaviour (18) criteria. In use case 1, Jev's most probable level fell short of the mean of the seven published judges by similar amounts on text-comparison and reasoning-heavy rubrics (difference in AC2 0.02, 95% CI −0.05 to 0.10), and its expected score by less on text-comparison rubrics (0.09, 0.03 to 0.14), a difference that disappeared when conciseness was left out (−0.04, −0.10 to 0.03; Fig. 3b); in use case 2 its shortfall from the agreement between physicians was larger on behaviour than on knowledge criteria (difference in macro F1 0.026, 95% CI 0.012 to 0.040; Fig. 3b, Supplementary Table 6; by theme, Extended Data Table 4).

## Discussion

Across four clinical evaluation tasks, a decision language model judged clinical AI output as well as, or better than, several LLM judges at a small fraction of their cost and time (Fig. 5). In agreement with physicians, Jev matched or exceeded seven of the 12 LLM judges in use case 1 with either reading, including Claude Opus 5.5 and GPT-6; in use case 2, it matched or exceeded four of six. In use case 3, it agreed with the MedHELM jury almost as well as the jurors agreed with each other, and more closely on completeness and, with its expected score, on accuracy; its expected score also agreed with clinicians as well as the jury and the best LLM judges. In use case 4, it agreed with GPT-5.4 on 90.8% of criteria, within 2.5 percentage points of every other judge.

Cost matters because evaluation recurs with every model update, prompt change and monitoring cycle, and token-priced frontier judges turn it into a large, unpredictable expense, a financial toxicity of adopting clinical AI[19,20]. Monitoring an assistant that answers 1,000 queries a day on seven ratings would cost about US$98 a year with Jev and US$12,400 to US$16,600 with Claude Opus 5.5 and GPT-6 at the prices billed (Fig. 5). In healthcare, routine evaluation of clinical generative AI output is becoming necessary for patient safety monitoring and governance[2,3], and a judge whose cost stays low and predictable as evaluation scales may help hospitals and other health organizations afford it.

Speed matters as well. With a few hundred milliseconds per request, Jev may allow the quality of a response to be checked in real time, whereas several seconds per judgment may be impractical for this purpose and better suited to retrospective audit. Unlike LLM judges, however, Jev returns only scores and their probabilities, without a rationale that could explain why a response failed and guide targeted improvements to prompts or the evaluated system.

Jev fell short in three places: on faithfulness, medical reasoning and actionability in use case 1 (Extended Data Table 2, Fig. 3c), on criteria below its confidence cutoff in use case 4 (Supplementary Table 4) and, with its most probable level, on the 56 clinician-rated responses of use case 3, a gap its expected score closed (Fig. 4a). These shortfalls have likely sources. The three rubrics of use case 1 require long context and slow, multi-step reasoning, matching Jev's documented difficulty with large inputs full of irrelevant detail and with multiple hops of reasoning[26]. Below the cutoff in use case 4, Jev's own confidence marks its judgments as uncertain, so they can be set aside or routed to another judge. On the 56 responses, the most probable level discards Jev's uncertainty between adjacent levels, which its expected score keeps.

Jev's developers recommend routing its low-confidence decisions to a human[27]; we tested routing them to an LLM judge instead[32]. LLM judges exceeded Jev substantially only in use case 1, where five agreed with physicians more closely, at 2.6 to 79 times Jev's cost, time or both (Table 1 and Extended Data Table 3, Supplementary Table 7), and hybrids added at most 0.04 to Jev's agreement in use cases 1 and 2, or 1.8 percentage points in use case 4, at 1.2 to 76 times Jev's cost (Fig. 5, Extended Data Fig. 2). An LLM judge or a hybrid may therefore be preferred only for specific evaluation tasks of high importance to stakeholders, such as safety screening, where routing only the safety rubric to Claude Opus 5.5 or GPT-6 raised detection of unsafe answers from 31% to 49% and 51% at a third of their cost; even then, expectations of the trade-off between cost and performance should be realistic.

A further contribution of this study is reading Jev's expected score, rather than its most probable level, as its rating. By keeping Jev's uncertainty between adjacent levels, the expected score raised the intraclass correlation with clinicians on the 56 clinician-rated MedHELM responses from 0.31 to 0.49 and raised its agreement with the MedHELM jurors on completeness to 0.87, against 0.79 between the jurors. Because both come from the same request, it adds no cost or time, and it may serve as a simple way to improve Jev in automated clinical evaluation tasks graded on ordinal scales.

Frontier judges agreed with physicians less closely than weaker or older judges in both use cases with physician labels: in use case 1, Claude Opus 5.5 and GPT-6 trailed Llama 4, GPT-4o and GLM-5.3 by 0.18 to 0.43 in AC2, and in use case 2 they ranked below every other judge, Jev included (Table 1, Supplementary Table 8). Frontier models judged fewer criteria met than physicians (46% and 41% against 77% in use case 2; Supplementary Table 8); whether this reflects overthinking the criteria, a divergence from how clinicians grade, or other causes was not tested here and warrants further study. Meanwhile, a judge should be chosen by its empirical measured agreement with clinicians rather than by its general capability.

Limitations of this study include the origin of use case 1 in a single institution and its grading in English translation (Jev's AC2 on the original Hebrew answers differed by at most 0.03; Supplementary Table 9). Use cases 2 to 4 use public data (use cases 2 and 3 with their labels) that Jev's undisclosed training may have contained. Use cases 3 and 4 measure agreement mainly with LLM graders. Three of the 13 MedHELM models also served as jurors and may have favoured their own responses[9], although Jev ranked the models in nearly the same order as the jury without them (Supplementary Table 3).

To conclude, a decision language model judged clinical AI output as well as, or better than, several LLM judges, mainly frontier models, across four tasks, with a median of 0.12 to 0.55 s per request; at the prices billed, frontier judges cost 73 to 294 times as much. While decision models are not meant to substitute LLMs in many clinical tasks that require broad medical knowledge and complicated reasoning, clinical AI output evaluation may be a particularly well-suited task for this type of model: many judgments can be structured around explicit criteria and context supplied in the prompt, while low cost and rapid responses could allow accuracy and safety checks to be run as often and as broadly as needed, including in real time.

## Methods

### Study design

This was a method-comparison study in which physician ratings were the reference standard in use cases 1 and 2. We analysed four use cases: answers from a pediatric clinical decision-support system that we had evaluated previously[16] (use case 1), the HealthBench meta-evaluation set[11] (use case 2), the open-ended MedHELM benchmarks[15] (use case 3) and HealthBench Professional[29] (use case 4). In use case 3, Jev, the MedHELM jury and the five LLM judges were compared with clinician ratings on the 56 clinician-rated responses, and Jev with the jury on all responses; in use case 4, which has no physician labels, Jev and the five LLM judges were compared with GPT-5.4, the benchmark's default grader. Jev graded every use case three times to measure stability. All runs took place from 13 to 18 October 2026. Reporting follows the TRIPOD-LLM[33] and MI-CLAIM-GEN[34] guidelines where applicable (TRIPOD-LLM checklist in Supplementary Data 1; MI-CLAIM-GEN checklist provided with the submission).

## LLM judges

Each use case was graded by LLM judges spanning frontier, low-cost and open-weight models (Extended Data Table 1). Five judges graded every use case: two frontier models, GPT-6 (OpenAI; gpt-6-astra) and Claude Opus 5.5 (Anthropic; claude-opus-5-5), and three open-weight models, MiMo-V2.6-Pro (Xiaomi), GLM-5.3 (Z.AI) and gpt-oss-120b (OpenAI). Each use case added its own: in use case 1, the seven judges of our earlier study, 12 judges in all, so that gpt-oss-120b appears twice, as the published run and as this study's run; in use case 2, GPT-5-nano (OpenAI; gpt-5-nano-2025-08-07), the least expensive current OpenAI model[35]; in use case 3, the published jury, with the five judges grading the 56 clinician-rated responses; and in use case 4, GPT-5.4, a frontier model and the benchmark's default grader and reference. Each LLM judge received its use case's prompt template unchanged and without labels: the original ordinal and binary prompts in use case 1 and the published grader templates in use cases 2 to 4; in use cases 2 and 4, outputs were read with the published HealthBench parser. These five judges graded use case 1 three times and use cases 2 to 4 once. In use case 2, Claude Opus 5.5 and GPT-6 graded a stratified half sample: half of the conversations in each of the seven themes (1,837 of 3,671 conversations, 14,775 rows), covering all 34 criteria (45% to 54% of each criterion's rows); GPT-5-nano and the open-weight models graded all rows. Providers, settings, re-requests and missing outputs are given in Supplementary Note 1.

## Use case 1: pediatric decision-support answers (published in npj Digital Medicine[16])

As detailed in the original study[16], a pediatric hospital's decision-support system answered 118 queries in Hebrew, each with a senior physician's reference answer, and five physicians graded each answer on six rubrics (faithfulness, actionability, medical reasoning, safety, conciseness and completeness) and a total score from 1 to 5 (4 or 5, pass). We reused the seven published judges' ratings and recomputed all statistics; Jev and this study's judges graded English translations of the answers from a re-run of the original translation step (gpt-4o-mini, temperature 0).

## Jev in use case 1

We used jev-1.13.0, version pinned, through the TypeSafe AI application programming interface[25]. For each answer, the input state held the query, the retrieved context and the English translations of the generated and reference answers. Jev evaluates each typed question about this state independently and returns a probability for each permitted answer.

For the ordinal scale, each request contained seven score questions, one per rubric and one for the total score, whose instructions reproduced verbatim the original prompt's role statement, scoring instructions, rubric definition and calibration examples; only the sentence naming the answer to grade, the list of inputs (now in the state), the instruction to score all metrics (now one metric) and the JSON output section were changed (Supplementary Note 2 gives every request). Levels 1, 3 and 5 carried the original anchors, and levels 2 and 4 were described as lying between their neighbours, with a short descriptor (Supplementary Note 2). Each rating was read as Jev's most probable level and as its expected score (Statistics and reproducibility). Score and yes or no questions were used rather than choice questions, for which the developers document a bias towards the first-listed option[26]. For the binary scale, each request contained six yes or no questions built from the original pass and fail definitions; an answer passed if the probability of yes was at least 0.5. Analyses used the first of the three runs. Jev was also run on the original Hebrew answers (Supplementary Table 9).

### Use case 2: physician-graded health conversations (HealthBench, a public benchmark)

In the HealthBench meta-evaluation set[11], each row pairs a multi-turn health conversation and a model response with one of 34 consensus criteria in seven themes, and two to five physicians recorded whether the criterion was met. We used the complete released file (version 2025-05-07, MIT licence): 29,511 rows from 3,671 conversations, with 60,896 labels from 186 physicians.

Jev received the conversation and the final response, without labels, and answered one yes or no question per criterion, whether the response met it. Each question reproduced the published HealthBench grader instructions with the criterion text[11], with the conversation moved to the state and the JSON-output sentences rewritten; criteria for the same response were sent in one request. A criterion was met if the probability of yes was at least 0.5, and the majority label of three runs was used.

### Use case 3: open-ended medical tasks graded by an LLM jury (MedHELM, a public benchmark)

MedHELM scores the open-ended responses of LLMs to medical tasks with a jury of three LLMs, GPT-4o, Claude 3.7 Sonnet and Llama 3.3 70B, each rating accuracy, completeness and clarity on a 5-point Likert scale with a benchmark-specific prompt that includes a reference response[15]. We used release v4.0.0, which publishes each response with its jurors' scores. Seven of its 14 jury-scored open-ended benchmark sets contain the response text: ACI-Bench, MEDIQA, MedicationQA, MedDialog (HealthCareMagic and iCliniq), MTSamples Procedures and MTSamples Replicate.

Their responses came from 13 models (Claude 3.5 Sonnet, Claude 3.7 Sonnet, DeepSeek-R1, Gemini 1.5 Pro, Gemini 2.0 Flash, Gemini 2.5 Pro, Llama 3.3 70B, GPT-4o, GPT-4o mini, GPT-5, GPT-5 mini, o3-mini and o4-mini). Jev graded the 45,385 of 45,461 responses with all nine jury scores; the Likert analysis excluded one more with an out-of-scale juror score (45,384).

Jev received, for each response, the blocks of the jury prompt (the input, the response and the reference response) and answered one score question per axis with five levels; the instructions reproduced the jury prompt's introduction (without its lines announcing the tagged blocks), its review instruction (with the tag brackets removed) and each axis definition (written on one line after the axis name), and levels 1 and 5 carried its anchors. Each rating was read in the same two ways as in use case 1.

The clinician ratings with which the authors validated the jury are released in the authors' analysis repository[36]: 56 responses from ACI-Bench (31) and MEDIQA (25), rated by 13 clinicians, one or two per response, all of them graded by Jev. Following the authors' analysis code[36], which reproduced their jury value of 0.47, each rater's score was the mean of the three axes, clinicians' scores were averaged per response, both columns were standardized, and agreement was the consistency intraclass correlation ICC(3,2), with 95% CIs from 2,000 bootstrap resamples of responses; the CI for the mean over clinician pairs came from resampling the pairs. A sensitivity analysis standardized each clinician's scores before averaging.

### Use case 4: real clinician tasks graded by GPT-5.4 (HealthBench Professional, a public benchmark)

HealthBench Professional (OpenAI, April 2026, MIT licence) comprises 525 tasks from physicians' conversations with ChatGPT for Clinicians (care consult, medical research, and writing and documentation; 191 deliberate red-teaming), each with physician-written criteria carrying points from −10 to 10 (1,135 criteria, 213 negative) and one physician-written response[29]. Its default grader is GPT-5.4 at low reasoning effort, and the release contains no grader outputs or physician labels. The LLM judges, including GPT-5.4 (gpt-5.4-2026-03-05, low reasoning effort), and Jev, with the use case 2 request, graded the 525 physician-written responses against their own criteria. Each grader received each criterion as "[points] criterion", one per request; for negative-point criteria, met means that the response shows the penalized behaviour.

## Rubric groupings

In use case 1, faithfulness, completeness and conciseness were text-comparison rubrics, gradable largely by comparing the answer with the context and reference answer, and actionability, medical reasoning and safety were reasoning-heavy rubrics, requiring clinical inference. In use case 2, the 34 criteria were classed as knowledge criteria (16, requiring a judgment of medical accuracy, safety or clinical appropriateness) or behaviour criteria (18, concerning context seeking, hedging, level of detail, tailoring to the user and instruction following).

## Statistics and reproducibility

### Two readings of Jev's ordinal scores

For every 1-to-5 score, Jev returns a probability for each level. We report two readings: its most probable level, with ties resolved to the lower level, Jev (most probable), and its expected score, Jev (expected), the sum of the levels weighted by their probabilities, as Jev returns it, plus 1 (Jev numbers the levels 0 to 4). Gwet's AC2, exact and within-one agreement, agreement between runs, the kept and deferred splits and the hybrids scored by AC2 used the expected score rounded half-up to the nearest level; Spearman and intraclass correlations, model ranking, mean bias and the hybrids scored by intraclass correlation used it unrounded, and the mean ratings of Supplementary Table 3 used it rounded. The deferral rule, which depends on Jev's score confidence, was the same for both readings. For yes or no questions, the expected value is the probability of yes, and rounding it gives the most probable label, so use cases 2 and 4 and the binary scale of use case 1 have a single reading. Both readings came from the same Jev outputs, analysis code, random seeds and bootstrap resamples.

### Agreement metrics

Because use case 1 ratings were concentrated at the top of the scale and at pass, where kappa-type coefficients, intraclass correlation and Krippendorff's alpha are unstable[37,38], we used, as in our earlier study, Gwet's AC2 with quadratic weights for ordinal ratings and Gwet's AC1 for binary ratings[39,40]. Each judge was compared with the physician panel, the median of the five physicians for each answer and rubric (rounded up when a missing rating left an even number). Agreement between physicians was the mean AC2 of each physician against the median of the other four, rounded up. Agreement between judges was the mean AC2 over pairs of judges (Supplementary Table 1). On the binary scale of use case 1, the reference was the majority of the five dichotomized physician ratings; with failure as the positive class, we computed AC1, sensitivity, specificity and positive and negative predictive values for Jev and this study's LLM judges from their ratings, and for four of the published judges from their published confusion matrices[16]. Bias was the judge's mean rating minus the panel's. Supplementary Table 8 adds exact and within-one agreement with the panel and the Spearman correlation with the physicians' mean rating, pooled over the seven ratings.

In use case 2, the primary metric was the macro-averaged F1 score of the HealthBench reference implementation[11]: every pair of a grader label and a physician label on a row is one comparison, and the F1 scores of the met and not-met classes are averaged. The physician reference applied the same metric to every ordered pair of physicians who labelled the same row, pooled over all pairs. Supplementary Table 8 adds accuracy, Cohen's kappa and sensitivity and specificity for not met.

In use case 3, for each axis, Jev's AC2 with each juror was averaged and compared with the mean AC2 of the three pairs of jurors; Spearman correlations across responses were averaged in the same way. Model ranking was the Spearman correlation across the 13 models between each model's mean Jev rating and its mean jury score, within each benchmark set, also without the three models that served as jurors[9]. In use case 4, with GPT-5.4 as the reference, we report the proportion of criteria with the same label, macro F1, AC1, sensitivity, specificity and predictive values (criterion not met as the positive class) and the correlation of task scores (points of met criteria divided by the sum of positive points, clipped to 0 to 1). Stability was the AC2 between runs (use cases 1 and 3) or the share of identical labels across three runs (use cases 2 and 4).

### Comparisons between judges

In use case 1, Jev was compared with each LLM judge, with the mean of the seven published LLM judges and with the mean of all 12 by the difference in AC2 with the physician panel, averaged over the seven ratings and for each rating; the three-run median and the average without conciseness, on which most judges disagreed with physicians, were sensitivity analyses. In use case 2, Jev was compared with each LLM judge, and every rater with the physician reference, by the difference in macro F1. A difference was called detectable when its two-sided 95% confidence interval excluded zero. Confidence intervals came from bootstrap resamples[41]: 2,000 of answers in use case 1, of conversations in use case 2 and of the 56 clinician-rated responses in use case 3, and 1,000 of responses in use case 3 otherwise and of criteria in use case 4; because the 13 MedHELM models answered the same inputs and HealthBench Professional criteria share tasks, the last two may be too narrow. All confidence intervals are percentile intervals; differences were recomputed within each resample, pairing the raters on the same items, with fixed seeds (2026 in use cases 1 and 2, 0 in use cases 3 and 4). Per-rubric, per-model, per-theme and rubric-group comparisons were not adjusted for multiplicity.

## Deferral and hybrid

To test deferral of low-confidence judgments, a judgment was kept if Jev's confidence reached a cutoff from 0 to 0.9 in steps of 0.1. Confidence was Jev's score confidence, returned with each score in use case 1 and recomputed in use case 3 from the stored probabilities with the developers' formula, $\max(0, 1 - \Sigma\, p_i|i - m|/1.2)$, m being the most probable level, and $|2p - 1|$ for a yes probability p in use case 2 (three-run mean) and use case 4 (first run). Kept and deferred judgments were compared at 0.5 for score confidence (use cases 1 and 3), the developers' documented floor for routing to a human[27], and at 0.4 for $|2p - 1|$ (use cases 2 and 4); the hybrids in Fig. 5 deferred judgments below 0.4. A cutoff changed no label; agreement was computed on the kept and deferred judgments, with the other raters on the same judgments. Because score confidence and $|2p - 1|$ are different measures, cutoffs are not comparable between them.

In the hybrid, Jev's first-run judgment was kept if its first-run confidence reached the cutoff and was otherwise replaced by the first-run rating of one of the use case's LLM judges (in use case 1, this study's five; in use cases 2 and 4, each LLM judge other than GPT-5.4, with Claude Opus 5.5 and GPT-6 on the half sample in use case 2; in use case 3, each juror, scored against the other two, and, on the 56 clinician-rated responses, each of the five LLM judges, axis by axis, scored by intraclass correlation with the clinicians). Binary use case 1 ratings used $|2p - 1|$ as confidence, and use case 3 used the confidence defined by Jev's developers, computed from the stored probabilities. In use case 2, the published HealthBench graders could not serve as fallbacks, because their row-level labels are not released. Hybrid cost was Jev's cost for all judgments plus the LLM judge's measured cost for the deferred ones (in use case 1, for every answer with any deferred judgment, since one request graded all rubrics; in use case 3, the estimated cost of one juror request per response with any deferred axis, averaged over the jurors, or, on the 56 clinician-rated responses, the LLM judge's measured cost per response with any deferred axis), on the same judgments as the LLM judge alone (use case 1, first ordinal run; use case 2, all 29,511 rows sent or the 14,762 half-sample rows compared). A hybrid's time per request was Jev's median time plus the LLM judge's multiplied by the deferred share (in use case 1, the share of answers with any deferred rubric). In use case 4, Jev and GPT-5.4 were timed at 16 concurrent requests and the five other LLM judges at six, so only GPT-5.4's time is compared with Jev's (Fig. 5). In the safety-only hybrid of use case 1, only the binary safety judgment was routed, at a cutoff of 0.2, which deferred a third of safety judgments (Jev's binary confidence runs lower than its ordinal confidence, so 0.4 would have deferred 58%); the outcome was sensitivity for answers the physician majority failed on safety, and cost was Jev's plus the LLM judge's binary request per deferred answer, relative to the LLM judge's binary cost.

### Cost and runtime

Billed input and output tokens were converted to cost at the October 2026 prices billed: batch prices for GPT-5-nano (all rows, although 10,021 ran synchronously), Claude Opus 5.5 and GPT-6 (apart from their synchronous rows) in use case 2, standard prices otherwise and, for the open-weight models, the cost reported by OpenRouter, including provider cache discounts[25,35] (standard list prices in Supplementary Table 7). The original study did not record token use, so we estimated the seven published judges' input tokens by tokenizing the original prompt with each answer (o200k tokenizer, approximate for the non-OpenAI models) at list prices (Supplementary Table 7)[35,42–44]; these costs exclude output and reasoning tokens and are lower bounds. Runtime was the client's end-to-end wall-clock time per synchronous request, from a cloud container in the central United States. Requests ran six at a time (Claude Opus 5.5's and GPT-6's six use case 1 runs in parallel; about 24 for Jev in use case 3; 16 for Jev and GPT-5.4 in use case 4). In use case 1, one request graded one answer on all rubrics, the LLM judges' times came from their three runs and Jev was timed in a separate pass of 118 ordinal requests; in use case 2, every judge but GPT-5-nano (36 of 100 sequential requests answered, the rest rate-limited) was timed on the same 150 random rows, and Jev also on 500 sequential requests. The MedHELM jury's cost, not published, was estimated by tokenizing its published prompts and juror outputs (o200k tokenizer) at list prices (GPT-4o at the gpt-4o rate, although the jury used the gpt-4o-2024-05-13 snapshot), with Amazon Bedrock's on-demand rate for Llama 3.3 70B[45] (Supplementary Table 7).

### Software

Analyses used Python 3.13 with NumPy 2.5, pandas 3.0 and SciPy 1.18, with custom implementations of AC1, AC2 and the HealthBench macro F1: AC1 reproduced the earlier study's published values from its confusion matrices, AC2 with quadratic weights reproduced Gwet's published two-rater example (0.9402)[40] and agreed with the irrCAC package for Python (version 0.4.4)[46] to within $10^{-12}$ on the use case 1 ratings of the physicians and published judges and the use case 3 ratings of Jev and the jurors, and macro F1 equalled the HealthBench reference functions.

### Ethics

Use case 1 was analysed under the waiver described in ref. 16: the Institutional Review Board (IRB) (Helsinki) Committee of Clalit Health Services granted a waiver of formal institutional approval and a waiver of informed consent (Reference Number: 0089-25-COM, issued on June 15, 2025). The queries were written by physicians and contain no patient data. Use case 2 is publicly released and anonymized[11]. Use cases 3 and 4 are publicly released[15,29]. No new human participants took part.

### Use of AI tools

During manuscript preparation and analysis, the authors used Claude (Anthropic) to assist with drafting and analysis. All AI-generated content was reviewed and validated by the corresponding author, who takes full responsibility for the content of the manuscript.

## Data availability

The HealthBench meta-evaluation data are openly available under the MIT licence from OpenAI (https://github.com/openai/simple-evals), as are the HealthBench Professional data (https://openaipublic.blob.core.windows.net/simple-evals/healthbench_professional/assets.zip). MedHELM release v4.0.0 responses and jury scores are openly available from the public bucket gs://crfm-helm-public/medhelm/benchmark_output/runs/v4.0.0/, and the MedHELM clinician ratings from https://github.com/som-shahlab/medhelm (file data/annotations/medhelm_human_annotation.csv, commit 8b0f3ef44f3aebdc27690fe9cc3867e90d3e0934). Physician and LLM ratings for use case 1 are available from the corresponding author on reasonable request. Jev and LLM judge outputs generated in this study are available at https://doi.org/10.5281/zenodo.23491933.

## Code availability

Code used to query Jev and the comparator models and to compute agreement statistics is available at https://doi.org/10.5281/zenodo.23491933.

## Funding

The authors declare no relevant funding.

## Author contributions

R.G. and N.D. conceived and designed the study. R.G., H.B.-Y., A.S., S.Y., O.T., N.G., M.R., A.A.H., I.A., and M.D. contributed to the development of the evaluation dataset, clinical rating, or implementation of the PANDA evaluation workflow. R.G. performed the evaluations of Jev and the LLM judges, designed and performed the statistical analyses, generated the figures and tables, and drafted the manuscript. S.C. contributed to the methodology, reviewed the manuscript and supervised the study. D.A. contributed to the methodology and reviewed the draft. R.G. and N.D. interpreted the findings. N.D. and R.B. supervised the study. All authors critically revised the manuscript, approved the final version, and agree to be accountable for the work.

## Competing interests

The authors declare no competing interests.

## Extended Data

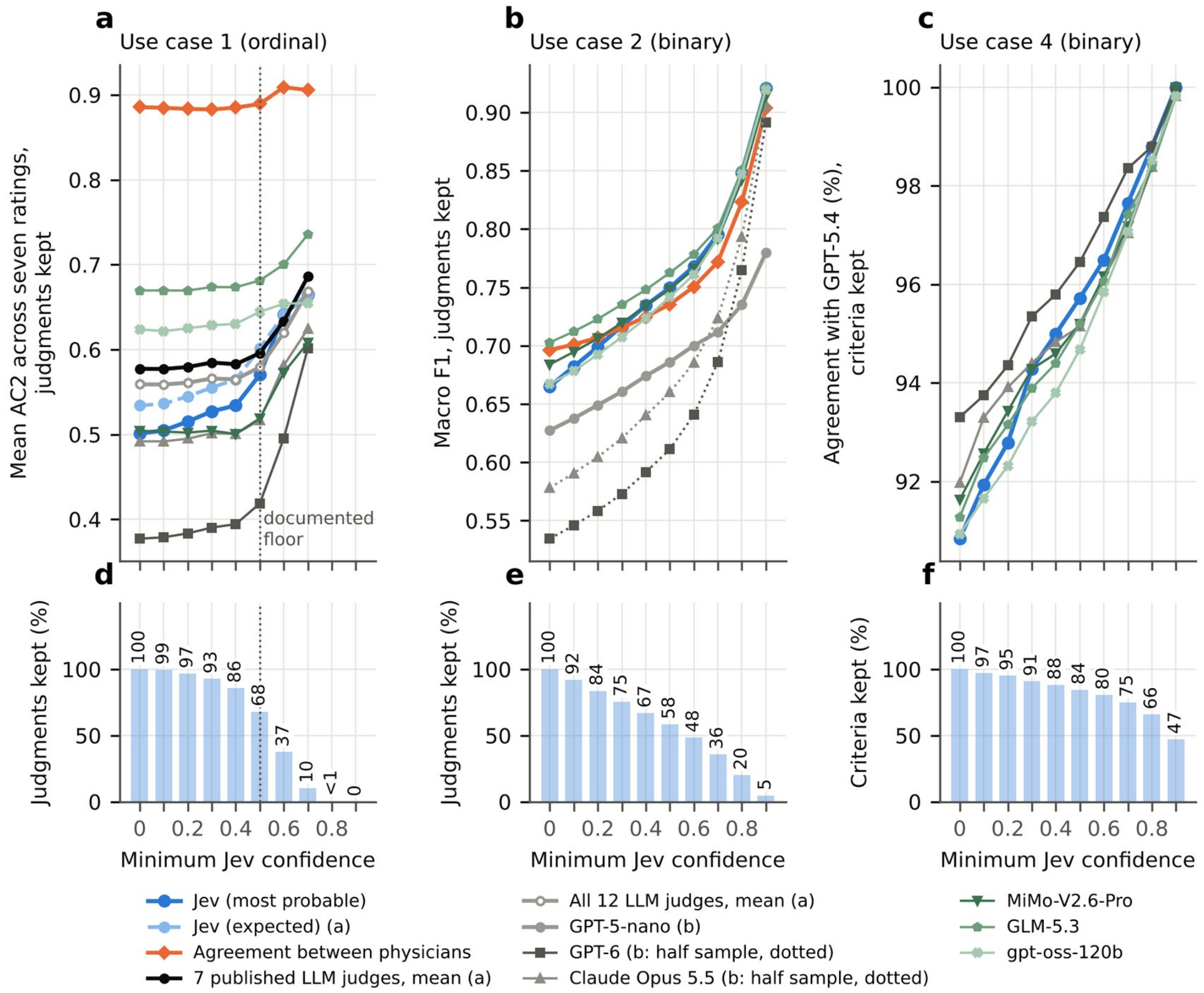

**Extended Data Fig. 1 | Agreement when Jev defers low-confidence judgments.** Each point keeps only the judgments at or above the minimum Jev confidence on the x-axis; the three use cases use the same 0.1 grid. In use case 1, confidence is the value Jev returns with each score; in use cases 2 and 4, it is $|2p - 1|$, where p is Jev's probability of yes, averaged over three runs in use case 2 and from the first run in use case 4 (0, an even split; 1, certainty). **a**, Use case 1: mean AC2 across seven ratings on the judgments kept for Jev (most probable), for physicians (each physician against the median of the other four), for Jev (expected) (dashed; Methods), for the mean of the seven published and of all 12 LLM judges, and for each of this study's five LLM judges; dotted line, the developers' documented floor of 0.5. Fewer than five answers remained above 0.7. **b**, Use case 2: macro F1 on the kept rows for Jev, for agreement between physicians, for GPT-5-nano and the three open-weight judges on all rows, and for Claude Opus 5.5 and GPT-6 on the half sample (dotted). **c**, Use case 4: agreement with GPT-5.4 on the kept criteria for Jev and the five other LLM judges (Claude Opus 5.5 on the kept criteria it labelled). **d** to **f**, Share of judgments (use case 1), rows (use case 2) or criteria (use case 4) kept at each cutoff.

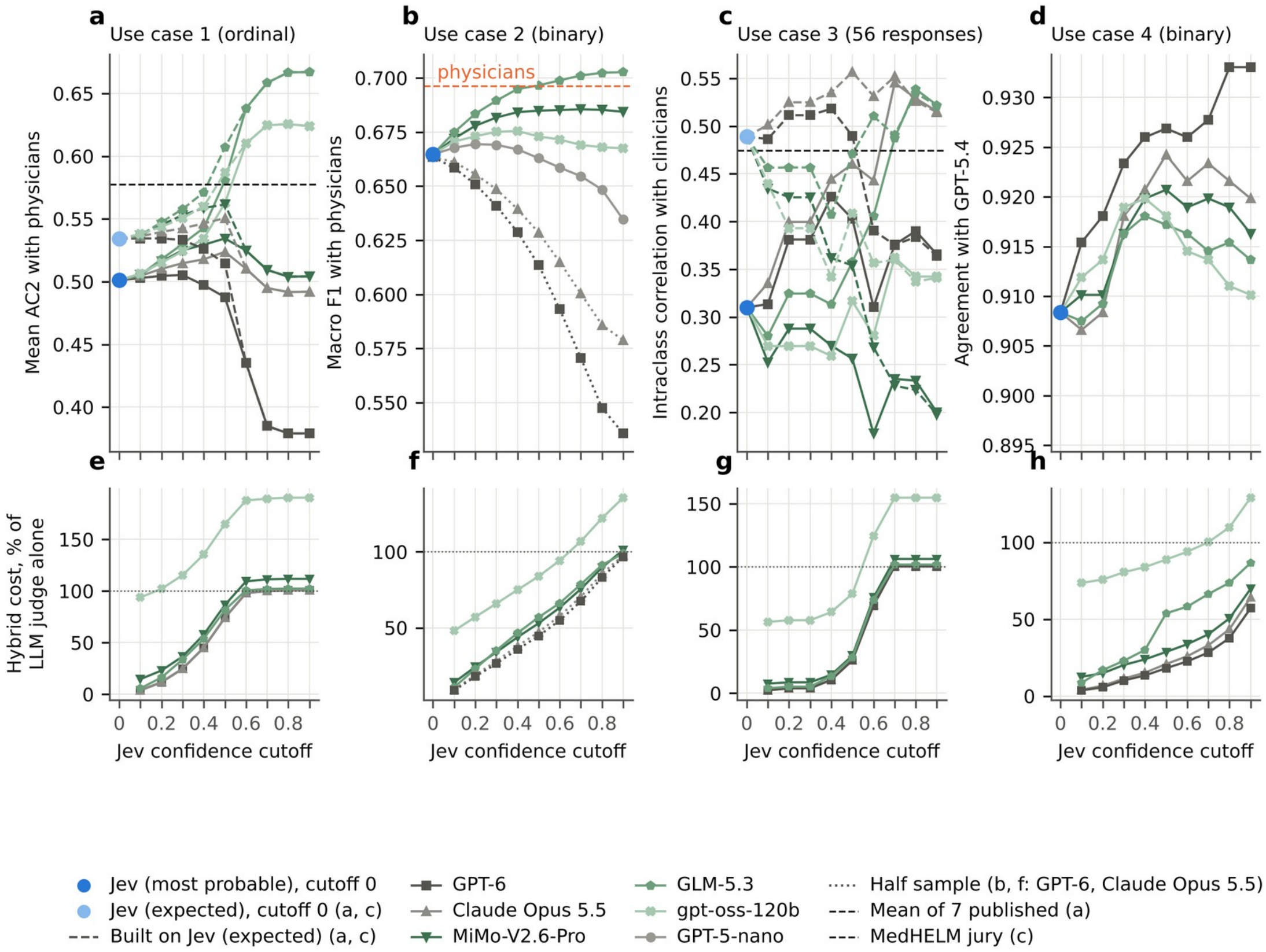

**Extended Data Fig. 2 | Agreement and cost of a hybrid of Jev and each LLM judge.** At each cutoff, Jev's judgments with confidence at or above the cutoff are kept and the rest are graded by the LLM judge (first runs); every judgment is scored against the use case's reference. Confidence as in Extended Data Fig. 1 (use case 3, the developers' score confidence; use case 4, $|2p - 1|$), but from a single Jev run; a cutoff of 0 is Jev alone (blue point). **a**, Use case 1: mean AC2 with the physician panel over the six rubrics and the total score, for the hybrid with GPT-6, Claude Opus 5.5, MiMo-V2.6-Pro, GLM-5.3 and gpt-oss-120b, built on Jev (most probable) (solid) or Jev (expected) (dashed; light blue point, Jev (expected) alone); black dashed line, mean of the seven published judges. **b**, Use case 2: macro F1 with physicians for the hybrid with GPT-5-nano and the three open-weight judges on all rows and with Claude Opus 5.5 and GPT-6 on the 14,762-row half sample (dotted); dashed line, agreement between physicians. **c**, Use case 3, 56 clinician-rated responses: intraclass correlation with clinicians when Jev's score on each axis below the cutoff is replaced by the score of each of the five LLM judges, for Jev (most probable) (solid) and Jev (expected) (dashed; light blue point, Jev (expected) alone); black dashed line, the MedHELM jury. **d**, Use case 4: agreement with GPT-5.4 for the hybrid with each of the five other LLM judges. **e** to **h**, The hybrid's cost in use cases 1 to 4 as a percentage of the LLM judge alone at the prices billed (GPT-5-nano omitted in f); dotted line, 100%. The hybrid's cost includes Jev on every judgment; in use cases 1 and 3, an answer or response with any deferred rubric or axis needs one full LLM request.

## Extended Data Table 1 | Use cases, raters and analysis

| | Use case 1: pediatric decision-support answers (published in npj Digital Medicine) | Use case 2: physician-graded health conversations (HealthBench, a public benchmark) | Use case 3: open-ended medical tasks graded by an LLM jury (MedHELM, a public benchmark) | Use case 4: real clinician tasks graded by GPT-5.4 (HealthBench Professional, a public benchmark) |
|---|---|---|---|---|
| Content | Answers of a deployed retrieval-augmented assistant to physician-written pediatric queries | Model responses to multi-turn health conversations | Responses of 13 LLMs to 6 open-ended benchmarks (7 benchmark sets): clinical notes, patient questions, medication questions, procedure reports | Physician-written responses to tasks physicians brought to ChatGPT for Clinicians (care consult, writing, research) |
| Language | Hebrew (judged in English translation) | English (mainly) | English | English (mainly) |
| Units rated | 118 answers | 29,511 response–criterion rows (3,671 conversations) | 45,384 responses | 1,135 response–criterion rows (525 tasks) |
| Rubrics | 6 rubrics + total score | 34 consensus criteria in 7 themes | Accuracy, completeness, clarity | Task-specific physician-written criteria (points −10 to 10) |
| Rating format | Ordinal 1 to 5 and binary pass or fail | Binary only (criterion met or not met) | Ordinal 1 to 5 | Binary only (criterion met or not met) |
| Physicians | 5 (30 of 4,130 ratings missing) | 186; 2 to 5 per row | 13 clinicians on 56 responses; none for the rest | None released |
| LLM judges | 12: 7 published (GPT-4o, o3, GPT-5, gpt-oss-120b, Gemini 3 Flash, Claude Sonnet 4.5, Llama 4), GPT-6, Claude Opus 5.5, MiMo-V2.6-Pro, GLM-5.3 and this study's run of gpt-oss-120b; 3 runs each | 6: GPT-5-nano, GPT-6[a], Claude Opus 5.5[a], MiMo-V2.6-Pro, GLM-5.3, gpt-oss-120b; 1 run | Published jury: GPT-4o, Claude 3.7 Sonnet, Llama 3.3 70B; on the 56 clinician-rated responses also GPT-6, Claude Opus 5.5, MiMo-V2.6-Pro, GLM-5.3, gpt-oss-120b; 1 run | 6: GPT-5.4 (default grader, low reasoning effort), GPT-6, Claude Opus 5.5, MiMo-V2.6-Pro, GLM-5.3, gpt-oss-120b; 1 run |
| Jev | jev-1.13.0, 3 runs | jev-1.13.0, 3 runs (majority label) | jev-1.13.0, 3 runs (first analysed) | jev-1.13.0, 3 runs (first analysed) |
| Reference | Median of 5 physicians (ordinal); majority of 5 after dichotomizing at 4 or above (binary) | Every ordered pair of physicians on the row, pooled | Each juror; jurors vs one another; clinicians (56 responses) | GPT-5.4 |
| Primary metric | Gwet's AC2, quadratic weights | Macro F1 (HealthBench definition) | Gwet's AC2, quadratic weights; model ranking (Spearman); consistency intraclass correlation ICC(3,2) with clinicians (56 responses) | Agreement, macro F1, AC1 |
| Run-to-run stability of Jev (3 runs) | AC2 between runs 0.99 for Jev (most probable) and 1.00 for Jev (expected) | Identical labels on 97.7% of rows | AC2 between runs 0.99 to 1.00 per axis (both readings) | Identical labels on 98.8% of criteria |
| Comparison | Jev and each LLM judge vs the physician panel | Jev and each LLM judge vs physicians | Jev vs jurors compared with jurors vs one another; Jev, the jury and each LLM judge vs clinicians (56 responses) | Jev and each LLM judge vs GPT-5.4 |

[a]Half sample stratified by theme, covering all 34 criteria (Methods). Jev (most probable) and Jev (expected), the two readings of Jev's scores (Methods). LLM, large language model.

## Extended Data Table 2 | Agreement of Jev and the 12 LLM judges with the physician panel in use case 1 (Gwet's AC2, ordinal scale)

| Rubric | Physicians, each vs other four | 7 published, range | 7 published, mean | GPT-6 | Claude Opus 5.5 | MiMo-V2.6-Pro | GLM-5.3 | gpt-oss-120b[a] | Jev (most probable) (95% CI) | Jev (expected) (95% CI) |
|---|---|---|---|---|---|---|---|---|---|---|
| **Mean of 7 ratings** | 0.89 | 0.41 to 0.81 | 0.58 | 0.38 | 0.49 | 0.50 | 0.67 | 0.62 | 0.50 (0.44 to 0.56) | 0.53 (0.49 to 0.58) |
| Faithfulness | 0.94 | 0.62 to 0.92 | 0.79 | 0.40 | 0.61 | 0.71 | 0.76 | 0.80 | 0.53 (0.39 to 0.67) | 0.58 (0.47 to 0.67) |
| Completeness | 0.86 | 0.61 to 0.90 | 0.79 | 0.65 | 0.73 | 0.70 | 0.83 | 0.80 | 0.77 (0.71 to 0.83) | 0.73 (0.67 to 0.79) |
| Conciseness | 0.90 | −0.95 to 0.60 | −0.30 | −0.70 | −0.24 | −0.32 | 0.05 | 0.07 | −0.25 (−0.40 to −0.10) | −0.05 (−0.13 to 0.04) |
| Actionability | 0.90 | 0.46 to 0.91 | 0.73 | 0.62 | 0.66 | 0.76 | 0.84 | 0.79 | 0.62 (0.51 to 0.71) | 0.60 (0.50 to 0.68) |
| Medical reasoning | 0.92 | 0.22 to 0.88 | 0.52 | 0.23 | 0.16 | 0.16 | 0.52 | 0.40 | 0.28 (0.12 to 0.42) | 0.31 (0.18 to 0.43) |
| Safety | 0.80 | 0.61 to 0.85 | 0.71 | 0.72 | 0.72 | 0.76 | 0.81 | 0.70 | 0.78 (0.71 to 0.84) | 0.78 (0.71 to 0.84) |
| Total score | 0.88 | 0.70 to 0.91 | 0.81 | 0.71 | 0.81 | 0.77 | 0.88 | 0.81 | 0.77 (0.69 to 0.84) | 0.79 (0.75 to 0.84) |

[a]This study's run. LLM and Jev ratings are from the first of three runs; confidence intervals from 2,000 bootstrap resamples of answers; each judge's own 95% CI, averaged over the seven ratings, is in Supplementary Table 8. Jev (most probable) minus each judge, averaged over the seven ratings (95% CI): Llama 4 −0.31 (−0.37 to −0.25), GPT-4o −0.23 (−0.29 to −0.17), GLM-5.3 −0.17 (−0.23 to −0.12), gpt-oss-120b, this study's run −0.12 (−0.18 to −0.07), Gemini 3 Flash −0.10 (−0.16 to −0.04), o3 −0.05 (−0.11 to 0.01), MiMo-V2.6-Pro 0.00 (−0.06 to 0.05), gpt-oss-120b, published run 0.00 (−0.07 to 0.08), Claude Opus 5.5 +0.01 (−0.05 to 0.07), GPT-5 +0.07 (0.004 to 0.12), Claude Sonnet 4.5 +0.09 (0.02 to 0.15) and GPT-6 +0.12 (0.06 to 0.19). Jev (most probable) minus the mean of the seven published judges was −0.08 (−0.12 to −0.03) and minus the mean of all 12 −0.06 (−0.11 to −0.02) (by rubric in Fig. 3c); with the median of three runs, −0.08 (−0.13 to −0.04) and −0.06 (−0.11 to −0.02); averaged over the six ratings other than conciseness, on which most judges agreed with physicians less than chance, −0.10 (−0.14 to −0.05) and −0.07 (−0.12 to −0.03). Jev (expected) minus each judge: Llama 4 −0.28 (−0.32 to −0.24), GPT-4o −0.20 (−0.25 to −0.14), GLM-5.3 −0.14 (−0.18 to −0.10), gpt-oss-120b, this study's run −0.09 (−0.14 to −0.04), Gemini 3 Flash −0.06 (−0.11 to −0.01), o3 −0.02 (−0.06 to 0.03), MiMo-V2.6-Pro +0.03 (−0.02 to 0.08), gpt-oss-120b, published run +0.03 (−0.04 to 0.11), Claude Opus 5.5 +0.04 (−0.01 to 0.09), GPT-5 +0.10 (0.05 to 0.15), Claude Sonnet 4.5 +0.12 (0.05 to 0.19) and GPT-6 +0.16 (0.09 to 0.22); minus the mean of the seven published judges −0.04 (−0.08 to −0.01) and of all 12 −0.02 (−0.06 to 0.01); with the median of three runs, −0.05 (−0.09 to −0.01) and −0.03 (−0.07 to 0.01); without conciseness, −0.09 (−0.13 to −0.06) and −0.07 (−0.10 to −0.03). Jev (most probable) and Jev (expected), the two readings of Jev's scores (Methods). Rubric-level comparisons are not adjusted for multiplicity. CI, confidence interval.

## Extended Data Table 3 | Cost and runtime of judging

| Judge | Work | Rubric judgments | Cost (US$) | Cost per million judgments (US$) | Time per request, median (10th to 90th percentile unless stated) | Full pass |
|---|---|---|---|---|---|---|
| Jev | Use case 1, both scales, 3 runs | 4,602 | 0.164 | – | 0.55 s (0.44 to 0.65; ordinal prompt, separate timing pass of 118 requests) | 11 s (6 parallel) |
| Claude Opus 5.5 | Use case 1, both scales, 3 runs | 4,602 | 22.79 | – | 5.72 s (4.42 to 8.39); maximum 13.8 s | Not recorded |
| GPT-6 | Use case 1, both scales, 3 runs | 4,602 | 29.82 | – | 6.07 s (4.61 to 8.13); maximum 14.2 s | ≈2 min (6 parallel) |
| MiMo-V2.6-Pro | Use case 1, both scales, 3 runs | 4,602 | 0.907 | – | 17.29 s | Not recorded |
| GLM-5.3 | Use case 1, both scales, 3 runs | 4,602 | 12.43 | – | 43.68 s | Not recorded |
| gpt-oss-120b, this study’s run | Use case 1, both scales, 3 runs | 4,602 | 0.199 | – | 18.33 s | Not recorded |
| 7 published LLM judges | Use case 1, ordinal, 3 runs (estimated) | 2,478 each | 0.048 (gpt-oss-120b) to 3.92 (Claude Sonnet 4.5), input only | 19.5 to 1,582, input only | Not recorded | Not recorded |
| Jev | Use case 1, ordinal, first run | 826 | 0.0317 | 38.4 | as above | as above |
| Claude Opus 5.5 | Use case 1, ordinal, first run | 826 | 4.01 | 4,854 | as above | as above |
| GPT-6 | Use case 1, ordinal, first run | 826 | 5.36 | 6,490 | as above | as above |
| MiMo-V2.6-Pro | Use case 1, ordinal, first run | 826 | 0.270 | 326 | as above | as above |
| GLM-5.3 | Use case 1, ordinal, first run | 826 | 1.34 | 1,624 | as above | as above |
| gpt-oss-120b, this study’s run | Use case 1, ordinal, first run | 826 | 0.0351 | 42.5 | as above | as above |
| Jev | Use case 2, one pass | 29,511 | 1.60 | 54.4 | 0.118 s (150 timing rows, six concurrent requests); 0.164 s (0.157 to 0.178, interquartile range; 500 sequential requests) | 80 to 83 s |
| GPT-5-nano | Use case 2, one pass | 29,511 | 8.06 (batch); 16.12 (standard); actual mixed run 10.74 | 273 (batch); 546 (standard) | 9.32 s (6.7 to 12.7, interquartile range; 36 requests) | ≈47 to 52 min (standard, extrapolated from 10,021 rows); batches 20 to 72 min each |
| Claude Opus 5.5 | Use case 2, half sample, one pass | 14,775 sent; 14,762 labelled and compared | 154.54; 153.63 for the 14,762 rows compared | 10,407 | 6.75 s | Not recorded |
| GPT-6 | Use case 2, half sample, one pass | 14,775 sent and labelled; 14,762 compared | 162.65; 162.48 for the 14,762 rows compared | 11,007 | 4.38 s | Not recorded |
| MiMo-V2.6-Pro | Use case 2, one pass | 29,511 | 35.82 | 1,214 | 13.29 s | Not recorded |
| GLM-5.3 | Use case 2, one pass | 29,511 | 260.80 | 8,837 | 22.41 s | Not recorded |
| gpt-oss-120b | Use case 2, one pass | 29,511 | 4.04 | 137 | 11.68 s | Not recorded |
| Jev | Use case 3, Likert scale, one pass | 136,155 (45,385 responses × 3 axes) | 2.96 | 21.7 | 0.12 s per response, all 3 axes (0.10 to 0.18) | Not recorded |

| Judge | Work | Rubric judgments | Cost (US$) | Cost per million judgments (US$) | Time per request, median (10th to 90th percentile unless stated) | Full pass |
|---|---|---|---|---|---|---|
| Jev | Use case 3, 56 clinician-rated responses | 168 | 0.00684 | 40.7 | 0.12 s per response, all 3 axes | as above |
| Claude Opus 5.5 | Use case 3, 56 clinician-rated responses, one pass | 168 (56 responses × 3 axes) | 1.85 | 10,986 | 9.96 s | Not recorded |
| GPT-6 | Use case 3, 56 clinician-rated responses, one pass | 168 | 2.01 | 11,982 | 5.75 s | Not recorded |
| MiMo-V2.6-Pro | Use case 3, 56 clinician-rated responses, one pass | 168 | 0.109 | 648 | 34.6 s | Not recorded |
| GLM-5.3 | Use case 3, 56 clinician-rated responses, one pass | 168 | 0.384 | 2,285 | 20.5 s | Not recorded |
| gpt-oss-120b | Use case 3, 56 clinician-rated responses, one pass | 168 | 0.0125 | 74.4 | 37.9 s | Not recorded |
| MedHELM jury (3 LLMs) | Use case 3, 56 clinician-rated responses | 168 (each by 3 jurors) | ≈1.18 (estimate; Supplementary Table 7) | ≈7,030 (estimate) | Not published | Not published |
| MedHELM jury (3 LLMs) | Use case 3 | 136,155 (each by 3 jurors) | ≈543 (estimate; Supplementary Table 7) | ≈3,989 (estimate) | Not published | Not published |
| Jev | Use case 4, one pass | 1,135 | 0.101 | 89.4 | 0.51 s (0.40 to 0.67) | Not recorded |
| GPT-5.4 | Use case 4, one pass | 1,135 | 7.44 | 6,556 | 3.52 s (2.52 to 6.15) | Not recorded |
| Claude Opus 5.5 | Use case 4, one pass | 1,135 sent; 1,121 labelled and compared | 19.35 | 17,047 | 3.14 s | Not recorded |
| GPT-6 | Use case 4, one pass | 1,135 | 23.93 | 21,083 | 2.94 s | Not recorded |
| MiMo-V2.6-Pro | Use case 4, one pass | 1,135 | 1.15 | 1,016 | 7.28 s | Not recorded |
| GLM-5.3 | Use case 4, one pass | 1,135 | 7.78 | 6,854 | 7.00 s | Not recorded |
| gpt-oss-120b | Use case 4, one pass | 1,135 | 0.144 | 127 | 9.89 s | Not recorded |

Time is end-to-end client wall-clock time from a cloud container in the central United States, including network transfer; in use case 1 one request graded one answer on all rubrics, and times are for the ordinal prompt (binary prompt medians: Jev 0.55 s, Claude Opus 5.5 6.12 s, GPT-6 6.01 s, slowest GPT-6 call 16.4 s). For MiMo-V2.6-Pro, GLM-5.3, gpt-oss-120b and, in use cases 2 and 4, Claude Opus 5.5 and GPT-6, the median is over the three ordinal runs (use case 1), 150 rows (use case 2) or all requests (use case 4), at six concurrent requests; the half-sample costs of Claude Opus 5.5 and GPT-6 are at batch prices apart from their synchronous rows. In use case 3, the five LLM judges were timed over all requests on the 56 clinician-rated responses at six concurrent requests, not comparably with Jev, so no multiple is given. Costs per million judgments are given on the basis of Fig. 5 (use case 1, first ordinal run; use case 2, all 29,511 rows sent or the 14,762 half-sample rows compared); the both-scales rows give the total spend. Multiples in the text and Fig. 5 use unrounded costs; billed tokens and list prices for every judge are in Supplementary Table 7. Jev bills input tokens only. Token use and timing for the seven published judges were not recorded in the original study; their input tokens were estimated by tokenizing the original prompt with each answer (o200k tokenizer), so their costs are lower bounds that exclude output and reasoning tokens. GPT-5-nano used about 1,480 input and 1,180 output tokens (about 1,060 of them reasoning tokens) per judgment. List prices in October 2026: Jev US$0.042 per million input tokens, output free; GPT-5-nano US$0.05 input and US$0.40 output per million tokens, half price through the Batch interface; GPT-5.4 US$2.50 input and US$15 output per million tokens[25,35]. In use case 4, Jev used 2.42 million input tokens and GPT-5.4 1.97 million input and 0.17 million output tokens; in use case 3, Jev used 70.4 million input tokens.

## Extended Data Table 4 | Agreement of Jev and the LLM judges with physicians in use case 2, by criteria group and theme (HealthBench meta-evaluation, macro F1)

| Criteria | Rows | Physicians | Jev | GPT-5-nano | MiMo-V2.6-Pro | GLM-5.3 | gpt-oss-120b | Rows[a] | Physicians[a] | Jev[a] | Claude Opus 5.5[a] | GPT-6[a] |
|---|---|---|---|---|---|---|---|---|---|---|---|---|
| **All criteria** | 29,511 | 0.696 | $0.665^{-}$ | $0.627^{-}$ | $0.684^{-}$ | 0.703 | $0.667^{-}$ | 14,762 | 0.686 | $0.664^{-}$ | $0.578^{-}$ | $0.535^{-}$ |
| Knowledge criteria | 13,883 | 0.674 | $0.658^{-}$ | $0.630^{-}$ | 0.669 | 0.682 | $0.646^{-}$ | 6,933 | 0.671 | 0.655 | $0.580^{-}$ | $0.523^{-}$ |
| Behaviour criteria | 15,628 | 0.712 | $0.670^{-}$ | $0.622^{-}$ | $0.694^{-}$ | 0.718 | $0.682^{-}$ | 7,829 | 0.696 | $0.669^{-}$ | $0.575^{-}$ | $0.544^{-}$ |
| Communication | 5,922 | 0.635 | $0.597^{-}$ | $0.553^{-}$ | $0.607^{-}$ | $0.617^{-}$ | $0.590^{-}$ | 2,964 | 0.644 | $0.607^{-}$ | $0.353^{-}$ | $0.320^{-}$ |
| Complex responses | 2,592 | 0.608 | $0.449^{-}$ | $0.533^{-}$ | $0.567^{-}$ | 0.581 | $0.537^{-}$ | 1,296 | 0.604 | $0.458^{-}$ | $0.436^{-}$ | $0.407^{-}$ |
| Context seeking | 3,230 | 0.700 | $0.734^{+}$ | $0.667^{-}$ | $0.728^{+}$ | $0.751^{+}$ | 0.683 | 1,615 | 0.677 | $0.724^{+}$ | 0.689 | $0.634^{-}$ |
| Emergency referrals | 3,612 | 0.689 | 0.692 | $0.661^{-}$ | 0.668 | 0.698 | $0.648^{-}$ | 1,808 | 0.675 | 0.696 | 0.654 | $0.575^{-}$ |
| Global health | 2,515 | 0.671 | $0.711^{+}$ | 0.687 | $0.711^{+}$ | $0.717^{+}$ | $0.697^{+}$ | 1,258 | 0.681 | $0.713^{+}$ | 0.692 | 0.651 |
| Health data tasks | 3,114 | 0.750 | 0.737 | $0.664^{-}$ | $0.695^{-}$ | 0.738 | $0.703^{-}$ | 1,564 | 0.760 | 0.736 | $0.616^{-}$ | $0.637^{-}$ |
| Hedging | 8,526 | 0.705 | $0.676^{-}$ | $0.620^{-}$ | $0.726^{+}$ | $0.738^{+}$ | 0.704 | 4,257 | 0.677 | 0.667 | $0.622^{-}$ | $0.554^{-}$ |

Macro F1 is the mean of the F1 scores for met and not met, computed over every pair of a rater label and a physician label on the same row; physicians, each physician against the others. $^{+}$ and $^{-}$, above or below agreement between physicians (95% confidence interval of the difference excludes 0; 2,000 bootstrap resamples of conversations). GLM-5.3, 29,507 rows. [a]Stratified half sample, graded by Claude Opus 5.5 and GPT-6. Jev, majority label of three runs; LLM judges graded once. Published graders on all rows[11]: GPT-4.1 0.709, o4-mini 0.692, o3 0.681, GPT-4.1 mini 0.661 and GPT-4.1 nano 0.580.